\pdfoutput=1
\documentclass[12pt,a4paper]{elsarticle}
\usepackage[top=2cm, bottom=2cm, left=2cm, right=2cm, footnotesep=1cm]{geometry}

\usepackage{amsmath,amsfonts}
\usepackage{booktabs}
\usepackage{cellspace}%
\usepackage{makecell}
\setcellgapes{3pt}

\usepackage{amssymb}

\usepackage{lineno}
\usepackage{graphicx}
\usepackage{subfigure}
\usepackage{booktabs}
\usepackage{algorithm}
\usepackage{algpseudocode}
\usepackage{amsmath}

\biboptions{authoryear,round}

\usepackage{url}

\journal{Journal}

\begin{document}

\begin{frontmatter}

\renewcommand\linenumberfont{\normalfont\scriptsize}


\title{Learning Spatiotemporal Features of Ride-sourcing Services with Fusion Convolutional Network}


\author[a]{Feng Xiao}
\author[a]{Dapeng Zhang}
\author[a]{Gang Kou}
\author[b]{Lu Li \corref{cor1}}
\cortext[cor1]{Corresponding author}
\ead{lliaw@connect.ust.hk}
\address[a]{School of Business Administration,Southwestern University of Finance and Economics, Chengdu, China.}

\address[b]{Business School, Sichuan University, Chengdu, PR China.}

\begin{abstract}

To collectively forecast the demand for ride-sourcing services in all regions of a city, the deep learning approaches have been applied with commendable results. However, the local statistical differences throughout the geographical layout of the city make the spatial stationarity assumption of the convolution invalid,  which limits the performance of CNNs on the demand forecasting task. In this paper, we propose a novel deep learning framework called LC-ST-FCN (locally connected spatiotemporal fully-convolutional neural network) to address the unique challenges of the region-level demand forecasting problem within one end-to-end architecture (E2E). We first employ the 3D convolutional layers to fuse the spatial and temporal information existed in the input and then feed the spatiotemporal features extracted by the 3D convolutional layers to the subsequent 2D convolutional layers. Afterward, the prediction value of each region is obtained by the locally connected convolutional layers which relax the parameter sharing scheme. We evaluate the proposed model on a real dataset from a ride-sourcing service platform (DiDiChuxing) and observe significant improvements compared with a bunch of baseline models. Besides, we also illustrate the effectiveness of our proposed model by visualizing how different types of convolutional layers transform their input and capture useful features. The visualization results show that fully convolutional architecture enables the model to better localize the related regions. And the locally connected layers play an important role in dealing with the local statistical differences and activating useful regions.

\end{abstract}

\begin{keyword}

3D convolution \sep Fusion \sep Locally-connected \sep Ride-sourcing \sep Spatiotemporal.



\end{keyword}

\end{frontmatter}


\section{Introduction}
\label{S:1}

In recent years, ride-sourcing service platforms, such as DiDiChuxing, Uber, and Lyft, have developed rapidly worldwide, aided by the growth of mobile internet, location-based services, cloud computing, and other innovative technologies. These platforms have various advantages compared to taxis, which suffer from temporal-spatial imbalance between supply and demand \citep{RN4}. Ideally, the unoccupied driving time of ride-sourcing services  is much less than that of taxis, which alleviates road congestion, reduces pollution emissions, and results in lower costs for the driver and mileage charges for the passenger. However, a new report from the \textit{San Francisco County Transportation Authority} showed that traffic congestion throughout the city increased from 2010 to 2016, where half of the congestion was attributed to the increase in ride hails \citep{RN62} due to unoccupied driving time, finding parking spaces, and locating passengers.

Although the efficiency of matching drivers and passengers has been greatly improved by ride-sourcing services \citep{RN1},  short-term prediction of passenger demand are still highly important for ride-sourcing service platforms so that available vehicles can be transferred from low-demand to high-demand areas in advance. This can effectively improve overall travel efficiency by reducing the waiting times for passengers, reducing ineffective mileage for drivers, increasing the matching rate, and decreasing costs.  Further optimization of the unoccupied driving time and accurate prediction of short-term travel demand are being aided by the development of electronic sensors and wireless communication technology, global positioning system  (GPS), global mobile communication system (GSM), and WiFi. At present, most ride-sourcing vehicles are equipped with such devices, which provide rich spatial and temporal information of the vehicle. These data have been very useful for supply and demand forecasting, fleet dispatching, travel time estimation, and route planning \citep{chen2017understanding,auer2017traffic}.

Various demand forecasting models have been developed up to date. Time series models are the most widely used methods, such as autoregressive integrated moving average model (ARIMA). Other machine learning methods have also been used, such as the deep neural network (DNN), the recurrent neural network (RNN), and convolutional neural network (CNN) models. Since the short-term demand forecasting models must consider both time and space, i.e., at a given time $t$, one must predict the demand in a certain region during the time period $[t,t+\Delta t]$, typical CNNs, including LeNet \citep{RN41}, AlexNet \citep{RN18}, and its deeper successors \citep{RN45, RN44}, cannot be used directly for demand forecasting problems. The detailed reasons are illustrated below:

(1) Time series statistics: In the case of demand forecasting problems, it is desirable to capture the time series statistics of demand for multiple adjacent time intervals. CNNs have been primarily applied to 2D feature maps to compute features only from spatial dimensions. Although typical CNNs can also take inputs with time dimension \citep{RN76} by arranging time series data in multiple channels, the temporal information is still collapsed completely after the first convolution layer.

 (2) Spatial coordinates: Typical CNNs ostensibly take fixed-sized inputs and produce non-spatial outputs as the fully connected layers have fixed dimensions and the spatial coordinates are lost \citep{RN17}. For short-term passenger demand forecasting, detailed predictions in all regions of a city depend on both global and local features, which need to be encoded using the spatial coordinates.

 (3) Parameter sharing scheme: By assuming spatial stationarity, the parameter sharing scheme can be used in convolutional layers, which dramatically reduce the number of parameters. However, the spatial stationarity assumption does not hold in a city where the local statistics or features vary from region to region.

In order to address these problems, we develop LC-ST-FCN, a new CNN-based DL model to handle the unique challenges of short-term passenger demand forecasting. In LC-ST-FCN model, the 3D convolutional operations are used to capture the time series statistics, which have achieved great performance on various video analysis tasks \citep{RN75, RN63}. Since 3D convolution is a natural extension of standard convolution, features from both spatial and temporal dimensions can be learned simultaneously. The overall architecture of the LC-ST-FCN is a fully convolutional network that takes input of arbitrary size and produces output of the same size; the spatial coordinates are maintained throughout the process and no spatial information is lost between layers. We fuse features across the layers to define a tunable nonlinear local-to-global-to-local representation where both global and local statistics are learned to improve the predictive performance.

For the demand forecasting problem of ride-sourcing services, which is essentially a regression problem, local differences in the statistics are critical for both the model structure and evaluation. We propose a weighted scheme to better compare different models. In addition, although the stationarity and randomness of data are seldom considered in DL models, we observed that the randomness arising from the data generating process led to learning difficulties in some regions. Hence, we classify all regions into two categories according to their randomness and evaluate the effect of uncertainty in the data on the predictive performance of DL models in demand forecasting problems.

The remainder of the paper is organized as follows. Section 2 reviews the existing literature. Section 3 describes the LC-ST-FCN model in detail. Section 4 outlines the evaluation of the model; we introduce the experimental dataset and metrics frequently used to evaluate model performance, analyze the inconsistency among those metrics and propose an improved method, and finally discuss the obtained results in detail. Section 5 concludes the study and proposes future research directions.

\section{Literature review}
\label{S:2}

Demand forecasting using spatial and temporal data collected by Internet and mobile terminals has becoming a research hotspot in the field of transportation.  \citet{RN21} used data from the community bicycle program \textit{Bicing} in Barcelona to analyze the pattern of human mobility in urban areas, and ARMA model was used to predict the number of bicycles available at different sites.  \citet{RN15} used the ARIMA model to forecast the demand of different taxi stations in Porto and Portugal, while the Markov algorithm, Lempel-Ziv-Welch algorithm, passion model, Moran's I values and others have been used to predict time-series data of traffic \citep{RN33, RN15, RN32}. Other studies used spatial clustering to mine taxi demand and GPS trajectory data, and studied the demand distribution of taxis in urban areas \citep{RN22, RN23}.

In recent years, DL technology has made tremendous progress, and has been widely used in many fields, including advanced speech recognition, visual object recognition, object detection, drug discovery, and genomics \citep{RN34}. DL can transform original information into a higher level with more abstract expression using a simple non-linear model. After sufficient combinations of transformations, very complex functions can be learned. Therefore, DL methods are being increasingly applied to traffic prediction problems.  \citet{RN35} proposed a neural network model composed of a deep belief network (DBN) and a multitask regression layer to predict short-term traffic flow. \citet{RN38} developed a depth-limited Boltzmann machine and RNN architecture to simulate and predict the evolution of traffic congestion. \citet{RN37} studied the travel behavior of ride-sourcing services using an ensemble method.

To realize accurate region-based forecasting, a number of deep learning models were designed to model the complex spatiotemporal information, and state-of-the-art results were achieved \citep{RN24, RN83, RN60, RN84, RN78}. Most of previous studies have focused on the combination of convolutional neural network and recurrent neural network, where the recurrent neural network architecture was utilized to model the temporal dependencies. The most related work to this paper is the model proposed by \citet{RN82}, in which the different components of temporal properties were learned separately and fused by a parametric-matrix-based method. Different from previous studies, the 3D convolution operations were employed to simultaneously capture the spatiotemporal dependencies in this paper. Other previous studies have also shown that 3D convolution operations have powerful ability to learn the spatiotemporal features \citep{RN73, RN72}. In particular, locally connected convolutional layers were used at the end of the model to obtain the final prediction results, without parameter sharing \citep{RN67, RN68, RN66}. To avoid the loss of spatial information, all of these convolution layers were applied with appropriate padding (only spatial) and stride 1, thus there was no change in term of spatial size from the input to the output of these convolution layers. Thus, this fully convolutional architecture was able to maintains the spatial coordinates of the input and no spatial information was lost between layers \citep{RN17}.


\section{The LC-ST-FCN model}
\label{S:3}

In this study, we partition a city into an $I \times J$  grid map based on the longitude and latitude, where a grid cell denotes a region. We divide the observation time period into $n$ time intervals with the interval length $\Delta t$ to produce set $T$, where $T={1, 2, ..., t, ..., n}$. During the period $[t,t+ \Delta t]$, the order demand generated in the region $(i, j)$ is $x_{t}^{(i,j)}$,  and the demand matrix of the entire $I\times J$ region is $X \in \mathbb{R}^{I \times J \times T}$.

\begin{figure}[ht]
    \centering\includegraphics[width=0.8\linewidth]{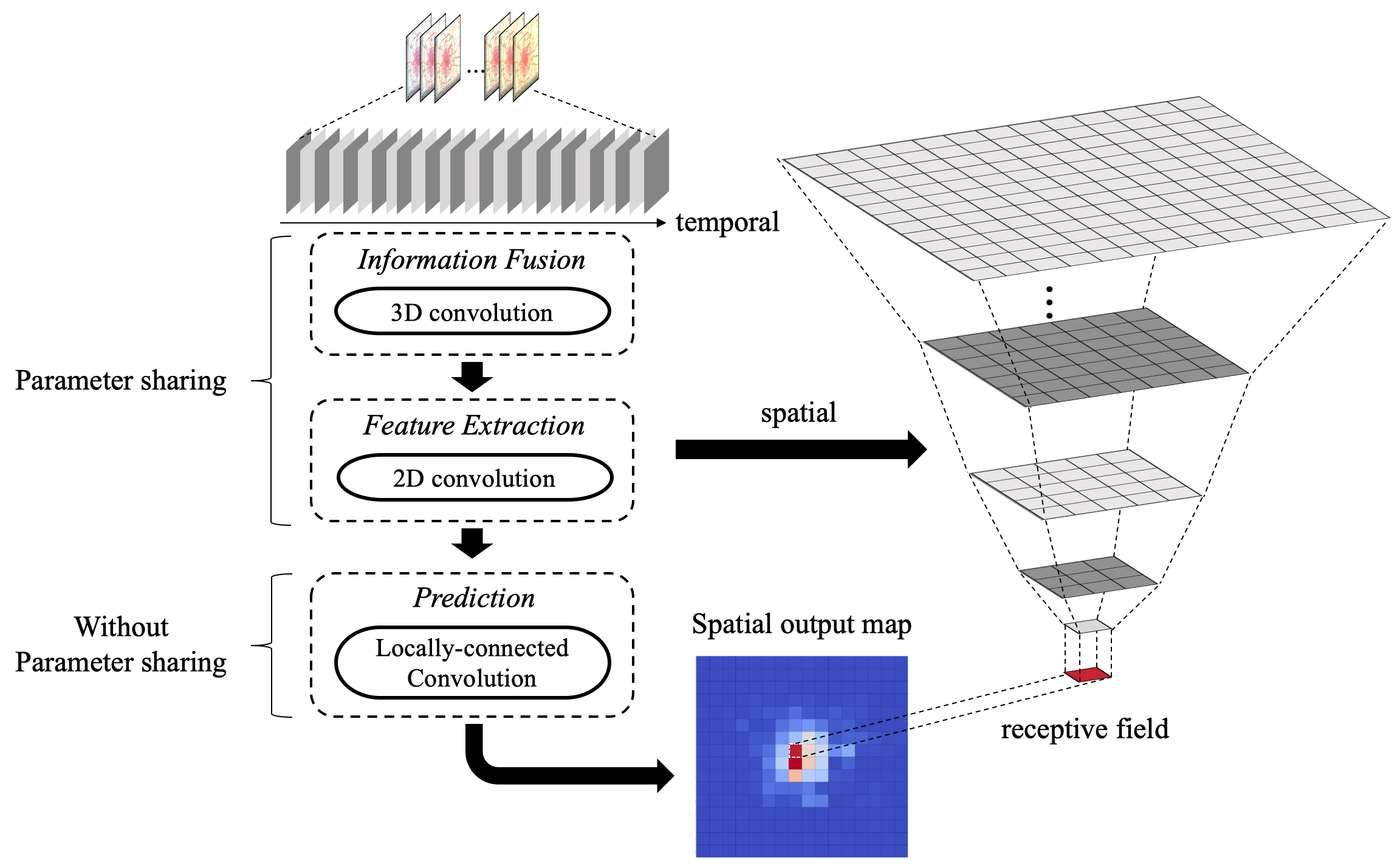}
    \caption{End-to-end implementation of the LC-ST-FCN model.}
    \label{End-to-end implementation of the LC-ST-FCN model.}
\end{figure}

Our goal is to predict how many orders will emerge during the future period $[t,t+ \Delta t]$ for each region at an instant $t$. That is, given the historical observations $\{X_t \mid t=0,...,n-1 \}$,  $X_{n}$ is predicted. The LC-ST-FCN model shown in Figure \ref{End-to-end implementation of the LC-ST-FCN model.} is developed to achieve this goal. We select a collection of short pieces from historical observations and stack them together to form a 3D volume as input. The 3D convolution operations are used to fusion the spatial and temporal information in multiple contiguous time windows. Since 3D convolution is a natural extension of standard convolution, features from both spatial and temporal dimensions can be learned simultaneously. Then, multiple 2D convolutional layers are used to extract and encode features from low to high level, and from local to global. With the increasing of network depth, the receptive field of neurons also increases, allowing an increasing amount of spatial information to be learned. In particular, locally connected convolutional layers are used at the end of the model to obtain the final prediction results, without parameter sharing.

According to the findings in 2D CNNs \citep{RN45}, small receptive fields of $3 \times 3$ convolution kernels with deeper architectures yield best results. Hence, in our LC-ST-FCN model we fix the spatial receptive field to $3 \times 3$ and vary only the temporal depth of the 3D convolution kernels. The 3D convolution kernel has access to information across all input demand matrix after several layers, depending on the depth of the input data. The  important components and training methods of the LC-ST-FCN model are introduced in the following sections.

\subsection{Input: 3D volume}

\begin{figure}[ht]
    \centering\includegraphics[width=0.8\linewidth]{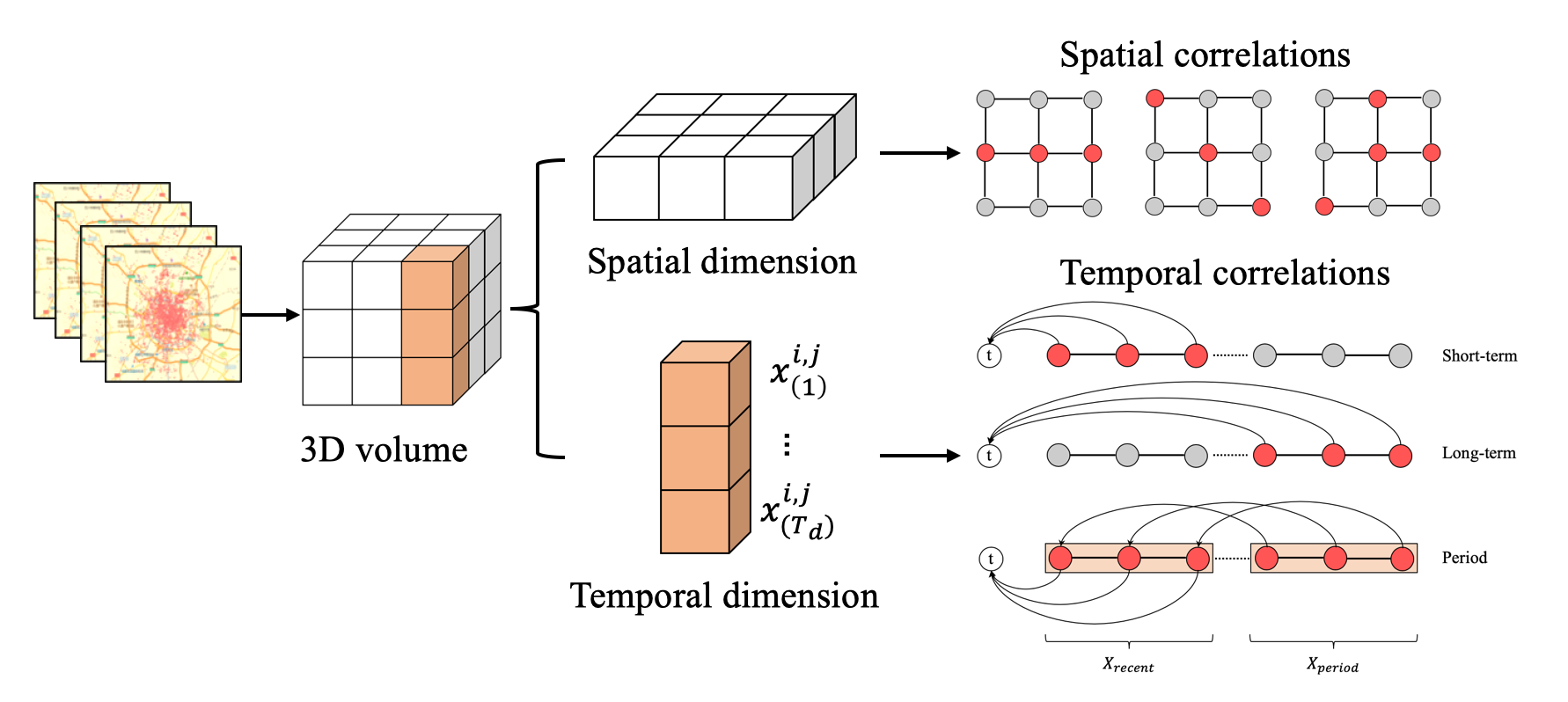}
    \caption{Schematic diagrams of the 3D volume of time series demand data.}
    \label{3D_tensor}
\end{figure}

As a class of attractive deep models for automated spatial feature construction, CNNs have been primarily applied on 2D images. However, for short-term passenger demand forecasting, it is also important to capture the temporal information in multiple adjacent or periodic time intervals. For example, there is a high demand for morning and evening rush-hour trips, but a low demand before dawn. Therefore, we construct the following input data forms, which integrate the spatiotemporal information of multiple time intervals into one 3D volume, with one demand matrix for each time interval:

    \begin{equation}
        \label{input_x}
        X^{input} = [X_{recent};X_{period}], X^{input} \in \mathbb{R}^{I \times J \times T_{d}}
    \end{equation}

    \begin{equation}
        \label{recent_x}
        X_{recent} = [X_{t-1},...,X_{t-T_{recent}}], X^{input} \in \mathbb{R}^{I \times J \times T_{recent}}
    \end{equation}

    \begin{equation}
        \label{period_x}
        X_{period} = [X_{t-L-1},...,X_{t-L-T_{period}}], X^{input} \in \mathbb{R}^{I \times J \times T_{period}}
    \end{equation}

    \begin{equation}
        \label{T_d}
        T_{d} = T_{recent} + T_{period}
    \end{equation}

Here, $T_{d}$, $T_{recent}$ and $T_{period}$ are the numbers of time intervals in $X^{input}$, $X_{recent}$ and $X_{period}$ separately. $L$ is the periodic length of the demand time series data. The periodicity of travel demand is the sharing feature of urban activities, so we use an additive model \citep{RN61} to determine the periodicity. Then, we decompose the overall travel demand $X_{t}^{all}$ in the city into trend term $X_{t}^{tre}$, periodic term $X_{t}^{per}$, and residual term $X_{t}^{res}$:

    \begin{equation}
        \label{X_t_all_1}
        X_{t}^{all} = \sum_{i}{\sum_{j}{x_{t}^{i,j}}}, t \in T
    \end{equation}

    \begin{equation}
        \label{X_t_all_2}
        X_{t}^{res}(L) = X_{t}^{all}-X_{t}^{tre}(L)-X_{t}^{per}(L) , 0 \le L \le T
    \end{equation}

    \begin{equation}
        \label{L_star}
        L^{*} = arg min {X_{t}^{res}(L)}
    \end{equation}

In the additive model, different $X_{t}^{res}$ values can be obtained by choosing different $L$ values, where  smaller $X_{t}^{res}$ values indicate that the corresponding $L$ is closer to the practical situation.
After $T_{recent}$, $T_{period}$ and $L$ are chosen, we can then construct the 3D volume, as shown in Figure \ref{3D_tensor}. Each instance used by the LC-ST-FCN model is a 3D volume containing $T_d$ number of raw samples with each sample represented by a matrix of $I \times J$ grid values.

\subsection{The fusion of fully connected 3D and 2D Convolutions}

\begin{figure}[ht]
  \centering
  \subfigure[2D convolution operations]{
    \label{2D}
    \includegraphics[width=0.4\linewidth]{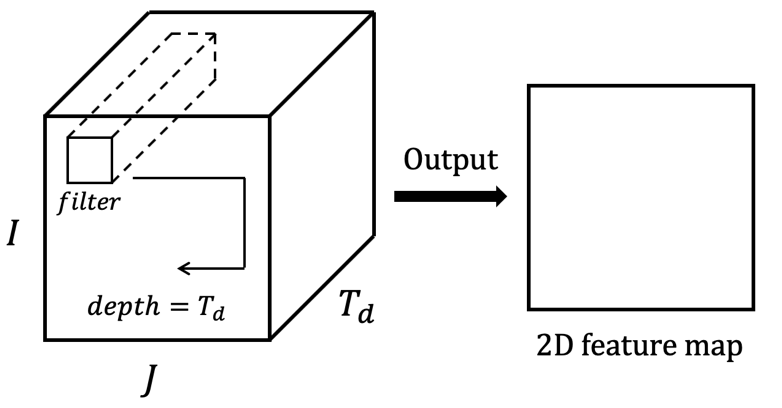}
  }
  \subfigure[3D convolution operations]{
    \label{3D}
    \includegraphics[width=0.4\linewidth]{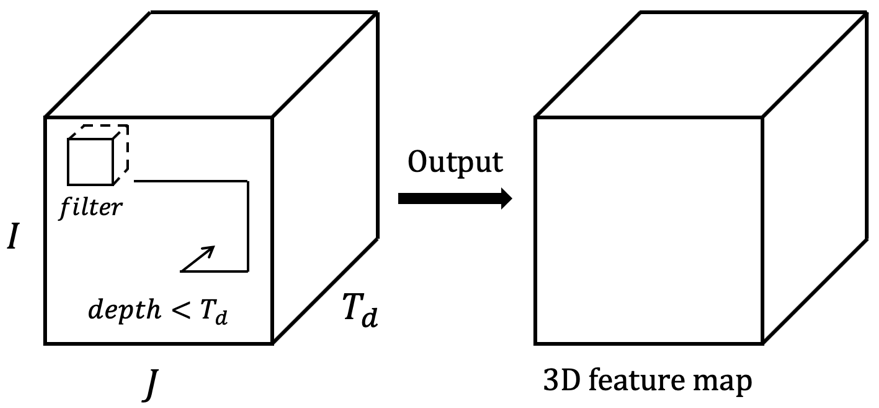}
  }
  \caption{The difference between 2D and 3D convolution operations.}
  \label{The difference between 2D and 3D convolution operations.}
\end{figure}

In 2D CNNs, convolutions are applied on the 2D feature maps to compute features from the spatial dimensions only. When the 2D CNN is applied to demand forecasting problem, it is desirable to capture the temporal information in multiple time intervals, e.g. trend and period. Although 2D CNN can also take multiple time intervals as input, after the first convolution layer, temporal information is collapsed completely. Hence, it is difficult to effectively extract temporal information by 2D convolution operations. Compared to 2D convolution, 3D convolution has the ability to model temporal information better, owing to that 3D convolution preserves the temporal information of the input by maintaining a 3D volume as the output. (As shown in Figure \ref{The difference between 2D and 3D convolution operations.}.)

Unlike 2D convolution operation, the depth of the 3D convolution filters is less than the depth of corresponding input volume. The feature maps obtained by the filter $k^{(c)}$ in the $c^{th}$ layer are given by:

    \begin{equation}
        \label{T_(c)}
        T_{d,k}^{(c)} = T_{d}^{(c-1)} - k_{d}^{(c)}+1
    \end{equation}

    \begin{equation}
        \label{X_k^(c)}
        X_{k}^{(c)} = ReLU(W_{k}^{(c)}*X^{(c-1)}+b_{k}^{(c)}), X_{k}^{(c)} \in \mathbb{R}^{I \times J \times T_{d,k}^{(c)}}
    \end{equation}

Where $W_{k}^{(c)}$ is the parameter matrices of filter $k$ connected to the feature maps in the previous layer, $k_{d}^{(c)}$ is the depth and $b_{k}^{(c)}$ is intercept parameter. The asterisk denotes the convolutional operator and $f$ is an activation function, convolving each filter across the width, height and depth of the input volume and computing dot products between the entries of the filter and input at any position. Each filter produces a separate 3D feature maps $X_{k}^{(c)}$, where the dimension of the depth of $X_{k}^{(c)}$ is $T_{d,k}^{(c)}$ obtained by valid convolution with stride 1 on previous layer with depth of $T_{d}^{(c-1)}$. Thus, finally the filter will be able to get access to all the information across the temporal dimension after a certain number of convolutions, depending on $T_d$. In the 3D convolutional layers, the filters have shared weights and the ReLU function is used \citep{RN18}.

After the 3D convolutional layers, we use a few 2D convolutional layers to further extract spatiotemporal features from low to high level and local to global. Stacking convolutional layers with tiny filters as opposed to having one convolutional layer with big filters allows us to express more powerful features of the input, and with fewer parameters.

\subsection{Locally connected convolutional layer}

The setting of parameter sharing in standard convolutional layers assumes local features. For example, if a horizontal boundary in some parts of the image is regarded important, then it will be equally useful in other places. However, the parameter sharing assumption may not be applicable for demand forecasting problems as the local features vary from region to region. For example, radial cities often have a typical central structure, and it is clearly inappropriate to use the same parameters to predict the demands for both the central and marginal areas of the city.

\begin{figure}[ht]
  \centering
  \subfigure[]{
    \label{loc_a.}
    \includegraphics[width=0.8\linewidth]{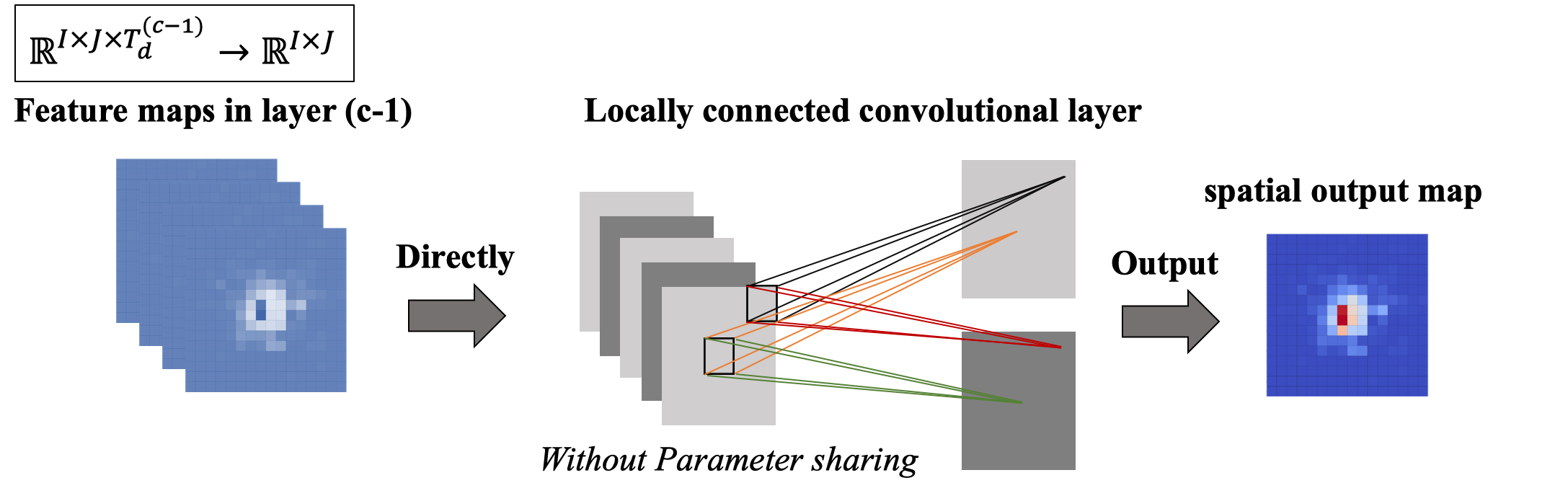}
  }
  \subfigure[]{
    \label{loc_b.}
    \includegraphics[width=0.8\linewidth]{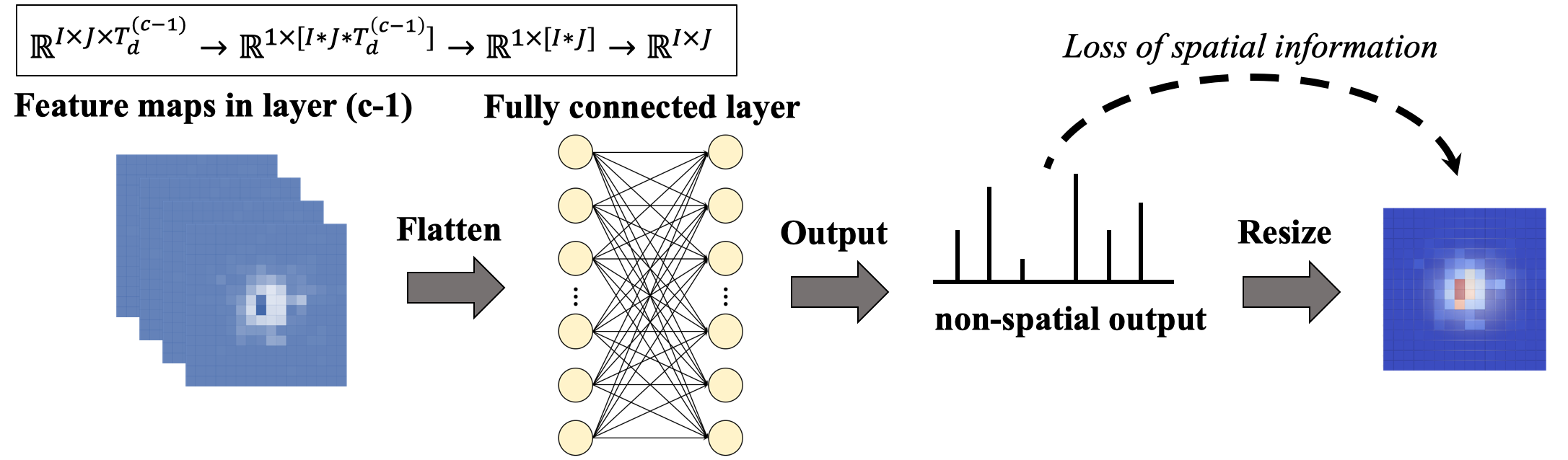}
  }
  \caption{Schematic diagram showing the transformation of fully connected layers into locally connected convolution layers to output a heatmap.}
  \label{locally_connected_layer}
\end{figure}

Hence, we relax the assumption of parameter sharing in standard convolution layers by using locally connected layers (without weight sharing) to obtain the final predictions. Like a standard convolutional layer, they apply a filter bank, except that every location in the feature maps is learnt by a different set of filters. As shown in Figure \ref{loc_a.}, in locally connected convolutional layers, the kernels in different spatial locations have different parameters. Compared with fully connected layers (Figure \ref{loc_b.}) or standard convolutional layers, locally connected convolution layers can simultaneously maintain local statistics and spatial coordinates.

\subsection{Objective function}

The LC-ST-FCN model can be trained by minimizing the mean squared error between the estimated demand $\hat{X}_{t}$ and real demand $X_t$. The objective function is shown in Eq.(\ref{loss}), where $w$ and $b$ are both learnable parameters. Algorithm 1 outlines the LC-ST-FCN training process, where the adaptive sub-gradient method is adopted for model training.

    \begin{equation}
        \label{loss}
         \mathop{\min}_{w,b} \ \ \|X_{t}-\hat{X}_{t} \|_{2}^{2}
    \end{equation}

  \begin{algorithm}[!ht]
  \caption{ LC-ST-FCN training algorithm.}
  \label{LC-ST-FCN training algorithm.}
  \begin{algorithmic}[1]
    \Require
      Historical demand of each region: ${X_0,X_1,...,X_n}$;
      Lengths of input data sequence: $T_d$;
      Lengths of recent and period sequence: $T_{recent}$, $T_{period}$;
      Lengths of period interval: $L$.
    \Ensure
      Learned LC-ST-FCN model.
    \State $//$ \textit{construct training instance:}
    \State $D \gets \emptyset$
    \For{all available time interval $t (1 \le t \le n)$}
      \State $X_{recent}=[X_{t-1},...,X_{t-T_{recent}}]$
      \State $X_{period}=[X_{t-L-1},...,X_{t-L-T_{period}}]$
      \State $//$ \textit{$X_t$ is the target at time $t$}
      \State put a training instance $(\{X_{recent},X_{period}\},X_t)$ into $D$
    \EndFor
    \State $//$ \textit{Training:}

    \Repeat
    \State Initialize the biases and weights at each layer;
    \State Sample minibatch from $D$ randomly;
    \If{in 2D or 3D convolution layers}
    \For{filters $k \in K$}
        \State $//$ \textit{Parameter sharing}
        \State {$k$ $\gets$ learnable parameters $w_k$,$b_k$}
    \EndFor
    \EndIf

    \If{in locally connected convolution layers}
    \For{filters $k \in K$}
        \State $//$ \textit{Without parameter sharing}
        \State {$k$ $\gets$ learnable parameters $w_k^{loc}$,$b_k^{loc}$}
    \EndFor
    \EndIf
    \State Calculate the stochastic gradient by minimizing the objective function
    \State shown in Eq.(\ref{loss});
    \State Update the parameters via back-propagation;
    \Until{\textit{stopping criteria are met}}

  \end{algorithmic}
  \end{algorithm}

\section{Experiments}
\label{S:4}

\begin{figure}[ht]
    \centering\includegraphics[width=0.8\linewidth]{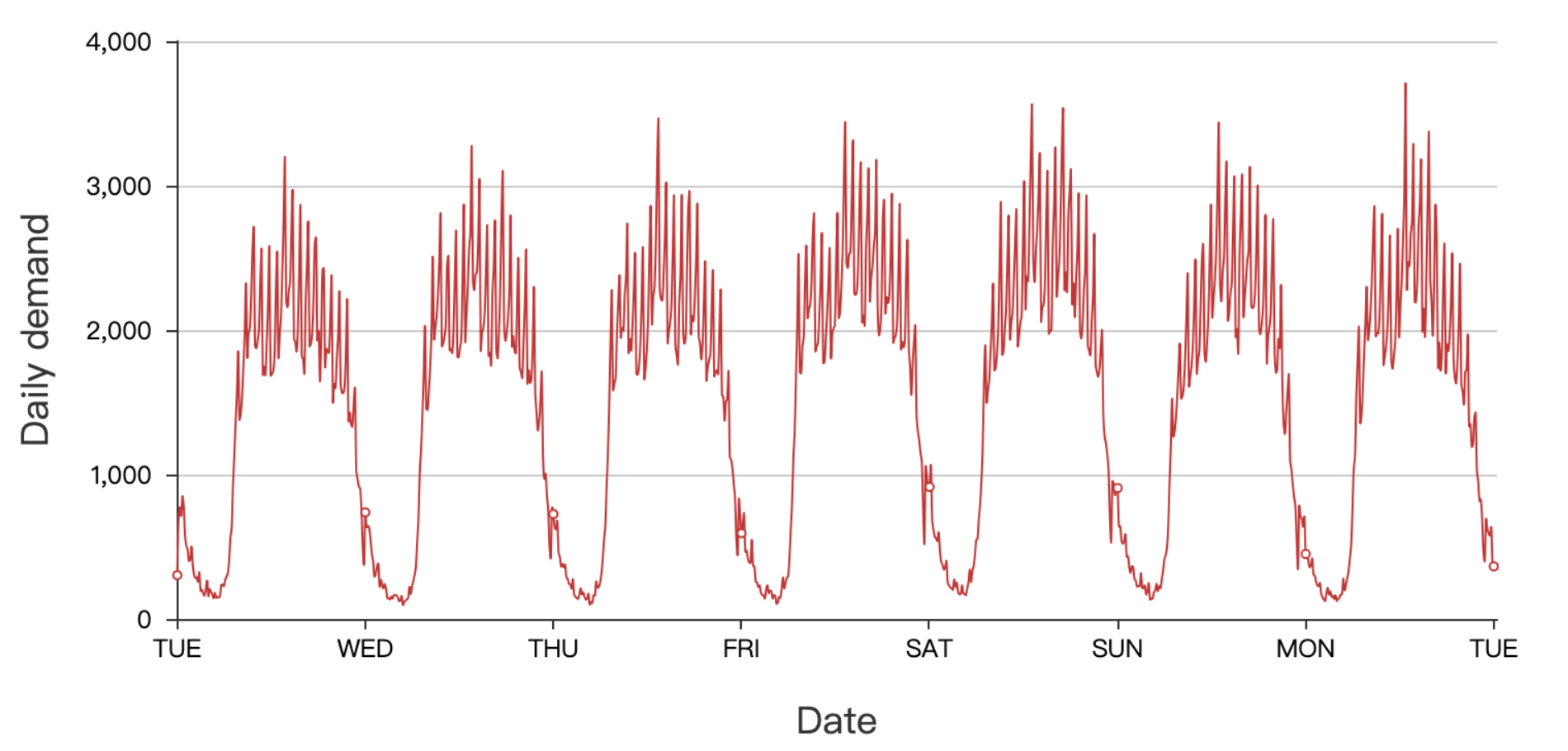}
    \caption{Daily demand in the first week of November 2016 (all regions).}
    \label{all_region_demand}
\end{figure}

We perform experiments using a dataset from DiDiChuxing \url{(https://gaia.didichuxing.com)}, which is the largest ride-sourcing service platform in China. The dataset includes customer requests in Chengdu, China, containing the request time, longitude, and latitude. After the raw data is cleaned, the dataset contained 7,031,022 requests within 103.85-104.30 longitude to the east and 30.48-30.87 latitude to the north are used in our experiments. All the requests are partitioned into 10-minute time intervals, and the investigated area is partitioned into $16 \times 16$ grids. Considering the periodicity of the data (see Figure \ref{all_region_demand}), we select the first three weeks of the dataset as a training set $D_{train}$, and the remaining nine days as an independent testing set $D_{test}$.

\begin{figure}[ht]
  \centering
  \subfigure[]{
    \label{trend.}
    \includegraphics[width=0.4\linewidth]{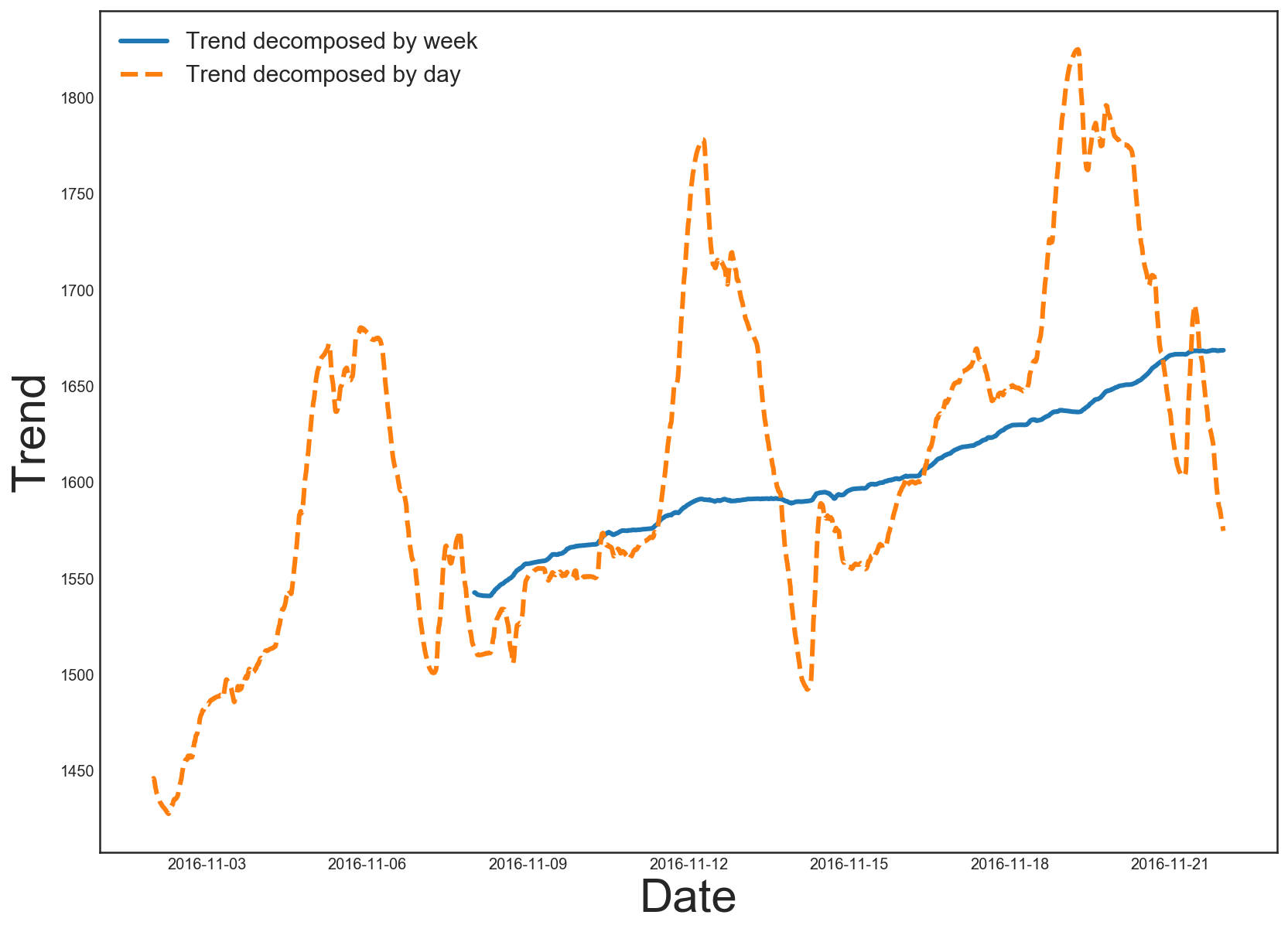}
  }
  \subfigure[]{
    \label{residual}
    \includegraphics[width=0.4\linewidth]{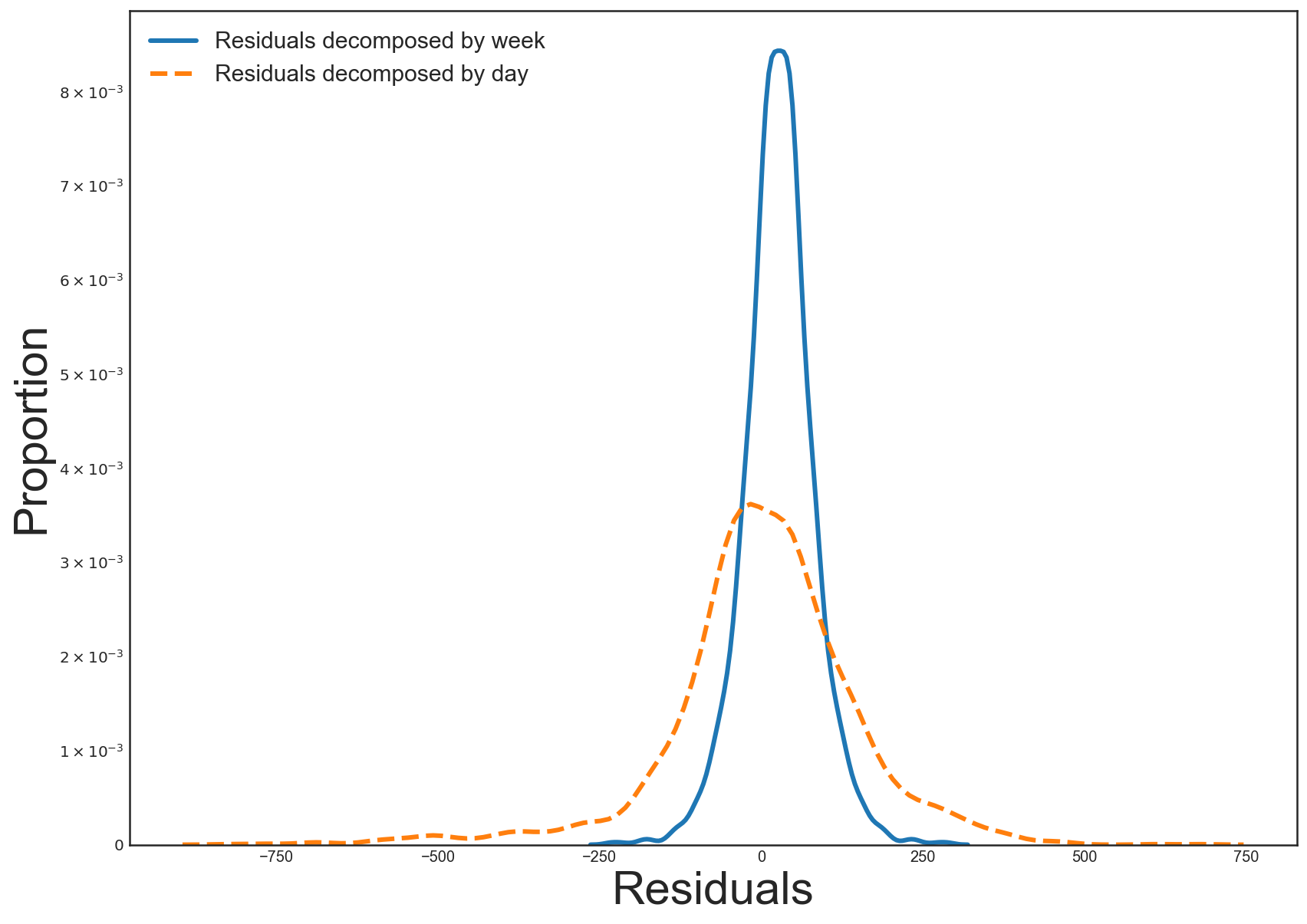}
  }
  \caption{Schematic diagram showing the transformation of fully connected layers into locally connected convolution layers to output a heatmap.}
  \label{Trend_and_residual}
\end{figure}

Because of the inherent periodicity of urban activity, before training we use the additive model to decompose the total demand data $X_t^{all}$ by time horizon length, $L$. After testing, we find that $L = 144$ (a day) or $1008$ (a week) yields better decomposition results than others. As shown in Figure \ref{Trend_and_residual}, for $L = 1008$,  $X_t^{tre}$ is almost linear and the distribution of residual terms has smaller variance than that for $L=144$. Hence, we select $L = 1008$ to construct our training 3D tensor for training. In order to illustrate the effectiveness of 3D convolution for learning time dependence (short-term and periodic), we compare our results with the traditional difference method \citep{RN61}. In the experiment, we set $T_{recent}= T_{period}=10$. Correspondingly, we use four 3D convolutional layers with filter depth $k_d=3,5,7,8$ to extract the spatiotemporal information of input. And the number of standard and locally-connected convolutional layers are 4 and 2.

To evaluate the performance of the LC-ST-FCN model, we select a bunch of different models as benchmarks, where all the models are trained and validated with the same training set and test set:

\begin{enumerate}[(1)]
\item LC-FCN: Variant of the LC-ST-FCN model, where the 2D convolutional layer are applied to the input to generate feature maps instead of 3D convolution layer. So, the temporal dimension of input is collapsed after the first 2D convolutional layer.

\item FCN: Variant of the LC-FCN model, where the last two layers are standard 2D convolutional layers (weight sharing), while the other parts of the structure remain the same.

\item LC-ST-FCN (diff): Variant of the LC-ST-FCN model, where the difference method is used to generate input data.

\item CNN: The last two layers are fully connected layers, while the other parts of the structure are the same as FCN.

\item Conv-LSTM: Conv-LSTM layers \citep{RN60} are used instead of convolution layers in FCN. Conv-LSTM structure can simultaneously learn spatial correlation and time dependence.

\item ANN (artificial neural network): Since different regions of a city have different local statistics, it is difficult to achieve effective prediction for all regions using a single ANN. Hence, we use a unique ANN for training and prediction in each grid region, resulting in a total of 256 ANNs that are trained. The input of 1-dimension time series data has a size of $1\times1\times20$, where 20 is the length of the temporal dimension.

\item Additive model: As for ANN, 256 independent addition models are used to obtain final prediction results for each region.

\item ARIMA: As ARIMA requires the stability and randomness of data, in our experiments, it is only used to generate predictions in regions that meet the requirements of stationarity and randomness.

\end{enumerate}

\subsection{Model comparison with arithmetic-mean-based metrics}

We first evaluate our model by the root mean squared error (RMSE), normalized root mean squared error (NRMSE), mean absolute percentage error (MAPE), and the modified MAPE methods according to a previous study \citep{RN51,RN15}, which are defined as follows:

\begin{equation}
    \label{RMSE}
    RMSE_{i,j} = \sqrt{\frac{1}{n}\sum_{r=1}^{n}(x_{t}^{i,j} - \hat x_{t}^{i,j})^2}
\end{equation}

\begin{equation}
    \label{NRMSE}
    NRMSE_{i,j} = \sqrt{\frac{\sum_{r=1}^{n}(x_{t}^{i,j} - \hat x_{t}^{i,j})^2}{\sum_{r=1}^{n}(x_{t}^{i,j})^2}}
\end{equation}

\begin{equation}
    \label{MAPE}
    MAPE_{i,j}= \frac{1}{n} \sum_{r=1}^{n} \frac{|x_{t}^{i,j} - \hat x_{t}^{i,j}|}{x_{t}^{i,j}+c}
\end{equation}

\begin{equation}
    \label{sMAPE1}
    sMAPE1_{i,j}= \frac{1}{n} \sum_{r=1}^{n} \frac{|x_{t}^{i,j} - \hat x_{t}^{i,j}|}{x_{t}^{i,j}+\hat x_{t}^{i,j}+c}
\end{equation}

\begin{equation}
    \label{sMAPE2}
    sMAPE2_{i,j}= \frac{\sum_{r=1}^{n}|x_{t}^{i,j} - \hat x_{t}^{i,j}|}{\sum_{r=1}^{n}|x_{t}^{i,j} + \hat x_{t}^{i,j}|}
\end{equation}

Here, $n$ is the number of all predicted values, $x_t^{i,j}$ and $\hat x_{t}^{i,j}$ are the ground truth and predicted value of demand in region $(i,j)$ at time interval $t$, respectively, while $c$ in Eq.(\ref{MAPE}) and (\ref{sMAPE1}) are set to 1.

Table \ref{unweighted} compares the predictive performances of the seven models on the test dataset. The numbers in the parentheses beside each value are the ranking of the predictive performance for the corresponding measurements. The results predicted by ARIMA are not included in the comparison range in Table 1 due to the requirement of stationarity and randomness of the data.

\begin{table}[!ht]
    \small
    \centering
    \caption{\newline Comparison of the predictive performance of the various models using the same test dataset.}
    \label{unweighted}
\begin{tabular}{@{}llllll@{}}
\toprule
Model            & RMSE              & NRMSE (\%)          & MAPE (\%)          & sMAPE1 (\%)        & sMAPE2 (\%)          \\ \midrule
LC-ST-FCN        & \underline{1.67 (1)} & 75.38 (2)           & \underline{22.40 (1)} & 13.31 (3)          & \underline{13.85 (1)} \\
LC-FCN           & 1.69 (2)          & \underline{75.28  (1)} & 22.98 (2)          & 13.45 (4)          & \underline{13.85 (1)} \\
FCN              & 1.78 (3)          & 99.67 (5)           & 27.36 (7)          & 15.91 (7)          & 16.93 (7)            \\
CNN              & 1.82 (4)          & 104.62 (6)          & 25.78 (6)          & 15.09 (6)          & 16.11 (5)          \\
Additive model   & 1.86 (5)          & 91.24 (4)           & 24.78 (4)          & 13.21 (2)          & 14.96 (4)            \\
Conv-LSTM        & 1.95 (6)          & 131.02 (8)          & 28.69 (8)          & 17.04 (8)          & 18.74 (8)            \\
ANN              & 2.04 (7)          & 76.92 (3)           & 24.43 (3)          & 14.23 (5)          & 14.53 (3)            \\
LC-ST-FCN (diff) & 2.17 (8)          & 105.53 (7)          & 25.72 (5)          & \underline{13.19 (1)} & 16.11 (5)          \\ \bottomrule
\end{tabular}
\end{table}

As shown by Table 1, LC-ST-FCN model obtains the best results on RMSE, MAPE and sMAPE2. Generally, LC-ST-FCN has better predictive performance than LC-FCN, which indicates that the 3D convolution operations capture the spatiotemporal information better than 2D ones. Compared with LC-FCN, FCN and CNN, the predictive performance of LC-FCN is significantly improved by simply replacing the last two standard convolutional layers or fully connected layers with locally connected convolutional layers. Which implies that the local-to-global-to-local architectures have excellent adaptability to the complex and changeable demand patterns of different regions in a city.

However, further comparison of the models is difficult due to the inconsistencies among the evaluation metrics. We observe the following interesting phenomena in Table \ref{unweighted}. The RMSE of CNN is less than that of the additive model, while all the relative errors (i.e. NRMSE and the others) are greater than that of the additive model. The same situation occurrs in the comparison of FCN and the additive model, CNN and ANN, FCN and ANN, and Conv-LSTM and ANN. Moreover, the contradictions among percentage error metrics are even more obvious. In general, relative errors are expressed as a ratio (unitless number), which can eliminate the influence of scale and better reflect the credibility of the measurement. However, when the absolute error and relative error are inconsistent in the evaluation results, it is difficult to effectively judge which model is superior. Such inconsistency is caused by the unbalanced demands of the regions. Thus, it is necessary to modify the evaluation criteria as well as the model structure, So that the evalutions of different models can be consistent.

\subsection{Issues in model evaluation}

Three issues will affect the effectiveness of the model evaluation: (1) the demand uncertainty; (2) the demand level; and (3) the demand distribution. We analyze each issue below.

(1) Demand uncertainty

\begin{figure}[ht]
    \centering\includegraphics[width=0.6\linewidth]{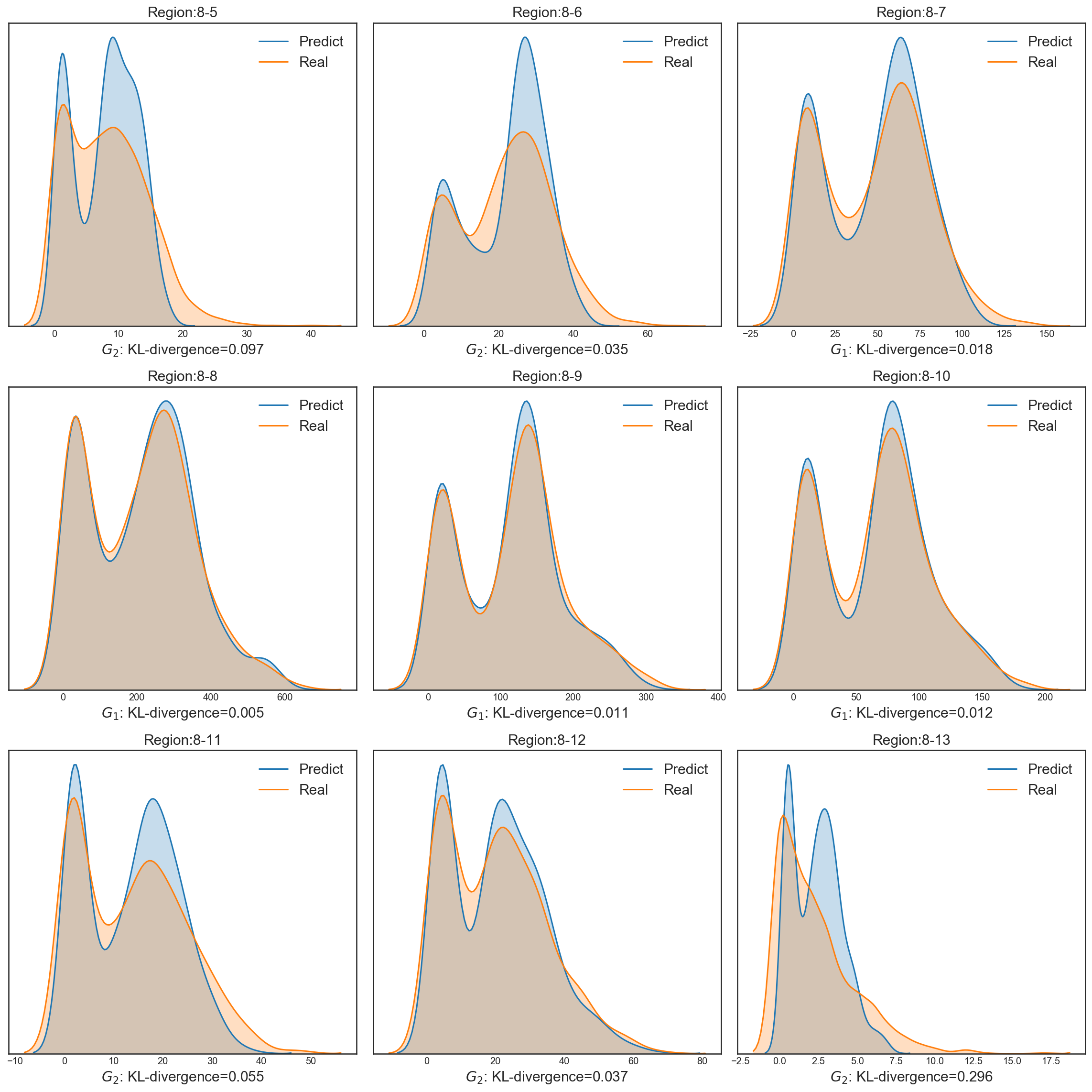}
    \caption{Frequency distribution of adjacent regions (from 8-5 to 8-13).}
    \label{KL_3D}
\end{figure}

In the literature, the uncertainty of data is seldom considered when training DL models. Especially, in short-term demand forecasting problems, such uncertainties significantly impact the model predictive performance. The CNN models do not output probability distributions. So attempting to extract the outcome of a sequence of a random time series data simply transfers the randomness from the inputs to the outputs. In each grid region, the historical observations $\{x_t^{i,j}\mid t=0,...,n-1;(i,j)\in(I,J)\}$ constitute an independent time series dataset. To extract the demand uncertainty, we use the \textit{Ljung–Box test} to assess the randomness of $x_t^{i,j}$. If the $p$ value of the test is greater than 0.05, $x_t^{i,j}$ is considered a random sequence. Based on the results of the \textit{Ljung–Box test}, the set of all regions, $G$, can be divided into a non-random sequence set $G_1$ and random sequence set $G_2$, where $G_1 \cap G_2=\emptyset$ and $G_1 \cup G_2=G$. In this case, for the total 256 grid regions, $G_1$ contains only 56 regions ($G_1/G=21.87 \%$),  but accounts for 80.88\% of the total demand, while $G_2$ contains 200 regions ($G_2/G=78.13\%$), but accounts for only 19.12\% of the total demand. As shown in Figure \ref{KL_3D}, we choose a group of regions adjacent to each other in latitude and acrossing the city center, and compare the frequency distributions of real and predicted data within the test data set. The difference between the two distributions is usually measured by the Kullback-Leibler (KL) divergence \citep{RN71}. We can easily observe that the gap between the two distributions is bigger for the regions in $G_2$ (regions 8-5, 8-6, 8-11, 8-12, 8-13) rather than  $G_1$, which implies that the model encounters greater difficulties in learning the distribution of $G_2$ regions with higher uncertainty. Therefore, it is necessary to treat the two different types of regions separately in the model evaluation.

(2) Demand level

\begin{figure}[ht]
  \centering
  \subfigure[]{
    \includegraphics[width=0.6\linewidth]{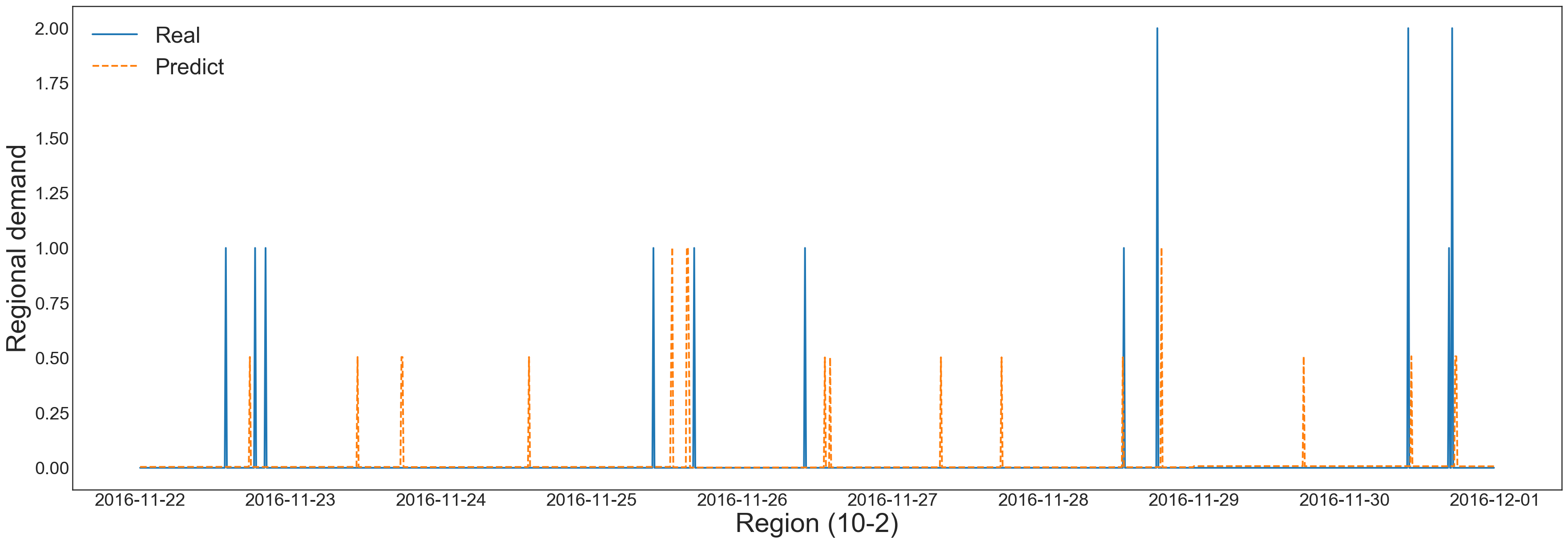}
    \label{102}
  }
  \subfigure[]{
    \includegraphics[width=0.6\linewidth]{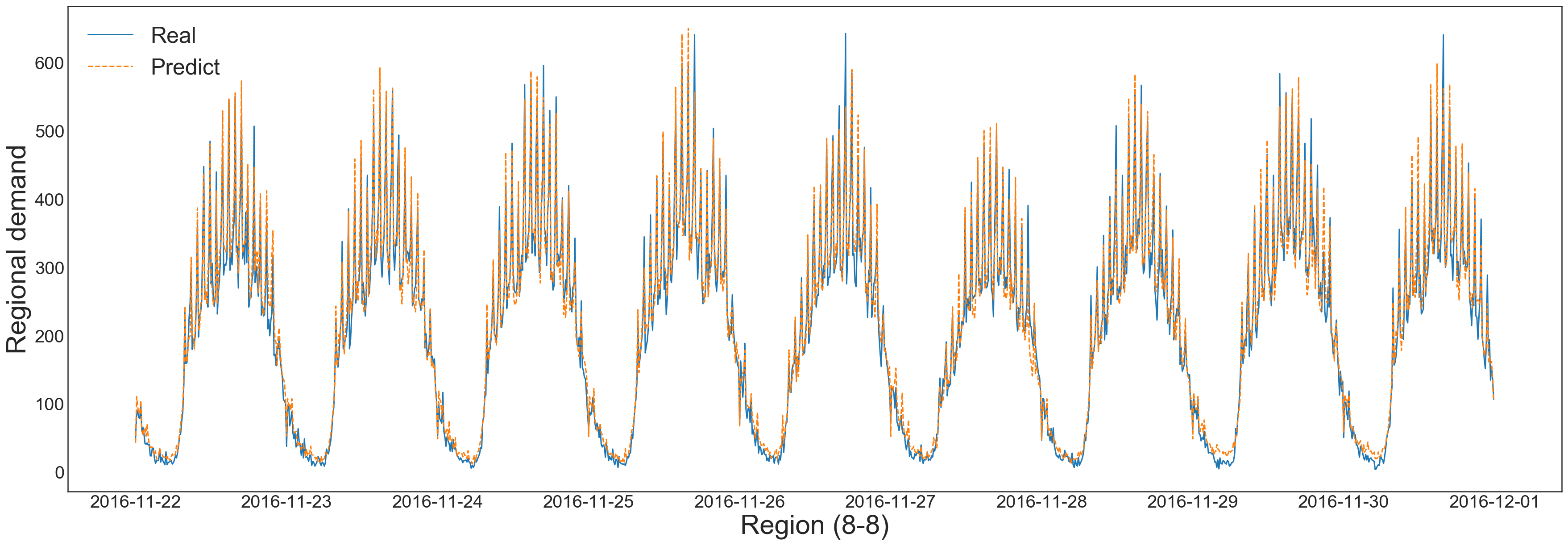}
    \label{88}
  }
  \caption{Prediction results for region 10-2 and 8-8 (Additive model).}
  \label{add_model} 
\end{figure}

The percentage error measurements should not be used to evaluate the overall performance of the model, since they are very sensitive to the level of absolute demand. For example, we compare the prediction results of the additive model for regions 10-2 and 8-8, as shown in Table \ref{Add_model} and Figure \ref{add_model}. In Table \ref{Add_model}, the model has a better performance for region 10-2 in almost all the percentage error measurements. However, Figure 10 indicates an opposite conclusion that the prediction for region 8-8 is better , since the pattern of demand oscillation is well learned by the model.

(3) Demand distribution

\begin{figure}[ht]
    \centering\includegraphics[width=0.6\linewidth]{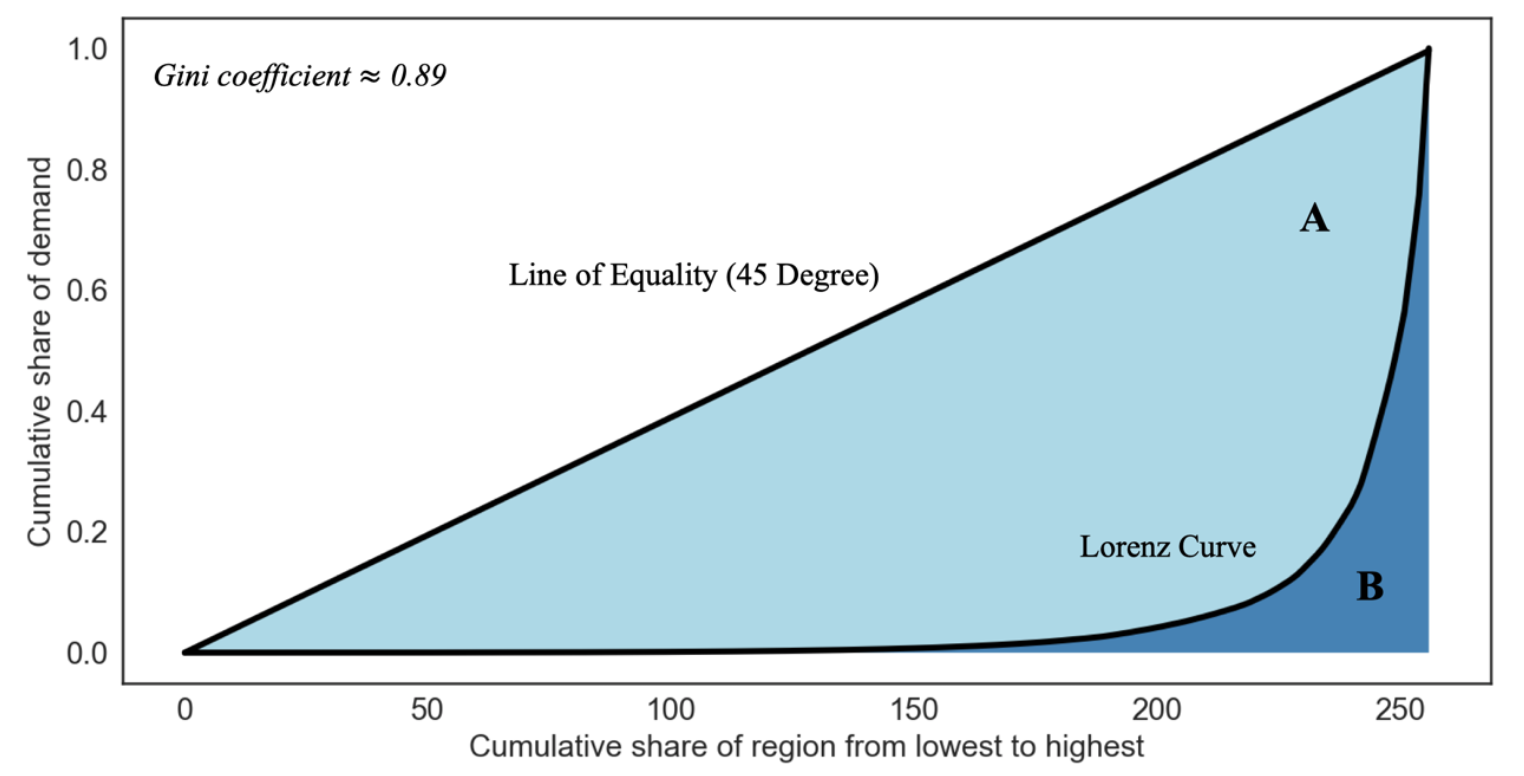}
    \caption{Gini coefficient of grid regions in the city.}
    \label{Gini}
\end{figure}

The Gini coefficient is most often used to measure how far a distribution deviates from a totally equal distribution. We calculate the Gini coefficient of the order demand distribution based on a Lorenz curve, which plots the proportion of the total demand of the city as a function of the cumulative share of the regional demand (as shown in Figure \ref{Gini}). The $y = x$ line represents perfect equality of demand. Hence, the Gini coefficient is the ratio of the area that lies between the lines of equality and the Lorenz curve (marked as A in the figure) to the total area under the line of equality (A + B in the figure). As a result of the variety of urban geography and layout, the Gini coefficient of our demand dataset is around 0.89, indicating extreme spatial unbalance of demand.

When the regional demand is highly unbalanced in a city, the evaluation results obtained by the arithmetic-mean-based methods are affected by the extreme values. The inherent averaging effect hides the complexity and heterogeneity behind the prediction results.

We sort all regions in increasing order in terms of demand level and calculate the cumulative moving average of the percentage errors of each model, as shown in Figure \ref{weighted}. We can observe that the curve is almost flat after the demand is larger than 25. This implies that the percentage measurement is determined mainly by the low demand regions, which occupy the majority of the prediction area. This may not be reasonable since the regions with large demand should be paid more attention.

From another point of view, when we increase the grid size from $16 \times 16$ to $64 \times 64$ with the same LC-ST-FCN structure, the prediction resolution is reduced. However, Table \ref{Grid1} shows a significant improvement for all the arithmetic-mean-based metrics, which indicates that the such measurements cannot objectively reflect the prediction ability of the model.

\begin{table}[ht]
    \small
    \centering
    \caption{\newline Prediction error of region 10-2 and 8-8 (Additive model).}
    \label{Add_model}
\begin{tabular}{@{}llllll@{}}
\toprule
                        & RMSE  & NRMSE (\%) & MAPE (\%) & sMAPE1 (\%) & sMAPE2 (\%) \\ \midrule
Region (10-2)  & 0.15  & 118.29     & 1.79      & 1.42        & 2.34        \\
Region (8-8)   & 23.06 & 9.13       & 17.90     & 7.27        & 4.08        \\ \bottomrule
\end{tabular}
\end{table}

\begin{table}[ht]
    \small
    \centering
    \caption{\newline Prediction errors of the LC-ST-FCN model with different grid sizes.}
    \label{Grid1}
\begin{tabular}{@{}llllll@{}}
\toprule
Grid size      & RMSE & NRMSE (\%) & MAPE (\%) & sMAPE1 (\%) & sMAPE2 (\%) \\ \midrule
$16\times16$      & 1.67 & 75.38      & 22.40     & 13.31       & 13.85       \\
$64\times64$      & 0.35 & 62.5       & 5.83      & 3.77        & 5.37        \\ \bottomrule
\end{tabular}
\end{table}

\subsection{Model comparison with weighted-arithmetic-based metrics}

In order to address the above issues, we propose weighted-arithmetic-based metrics instead of the widely used arithmetic-mean-based metrics. The weight $\alpha_{ij}$ is obtained by calculating the ratio of the regional demand to total demand for the training set $D_{train}$. All the new metrics are shown in Table \ref{The weighted-arithmetic-based metrics.}.
\begin{equation}
    \label{weight_coef}
    \alpha_{ij}  = \frac{\sum_{t \in D_{train}}{x_{t}^{i,j}}}{{\sum_{t \in D_{train}}\sum_{i,j \in I,J }}x_{t}^{i,j}}
\end{equation}

\begin{table}[ht]
  \centering
  \footnotesize
  \caption{The weighted-arithmetic-based metrics.}
  \label{The weighted-arithmetic-based metrics.}
  \begin{tabular}{@{}cc@{}}
    \toprule
    \textbf{Arithmetic-mean-based metrics} & \textbf{Weighted-arithmetic-based metrics} \\ \midrule
   $NRMSE = \displaystyle \frac{1}{I  J}\sum_{i,j \in I,J}{NRMSE_{i,j}} $      &    $W_{-}NRMSE = \displaystyle \sum_{i,j \in I,J}{\alpha_{ij}NRMSE_{i,j}} $ \\\addlinespace%

   $MAPE = \displaystyle \frac{1}{I  J}\sum_{i,j \in I,J}{MAPE_{i,j}} $          &  $W_{-}MAPE = \displaystyle \sum_{i,j \in I,J}{\alpha_{ij}MAPE_{i,j}} $ \\\addlinespace%

   $sMAPE1 = \displaystyle \frac{1}{I  J}\sum_{i,j \in I,J}{MAPE1_{i,j}} $        &  $W_{-}sMAPE1 = \displaystyle \sum_{i,j \in I,J}{\alpha_{ij}sMAPE1_{i,j}} $ \\\addlinespace%

    $sMAPE2 = \displaystyle \frac{1}{I  J}\sum_{i,j \in I,J}{sMAPE2_{i,j}} $  &   $W_{-}sMAPE2 = \displaystyle \sum_{i,j \in I,J}{\alpha_{ij}sMAPE2_{i,j}} $ \\ \bottomrule
  \end{tabular}
\end{table}

\begin{figure}[ht]
    \centering\includegraphics[width=0.6\linewidth]{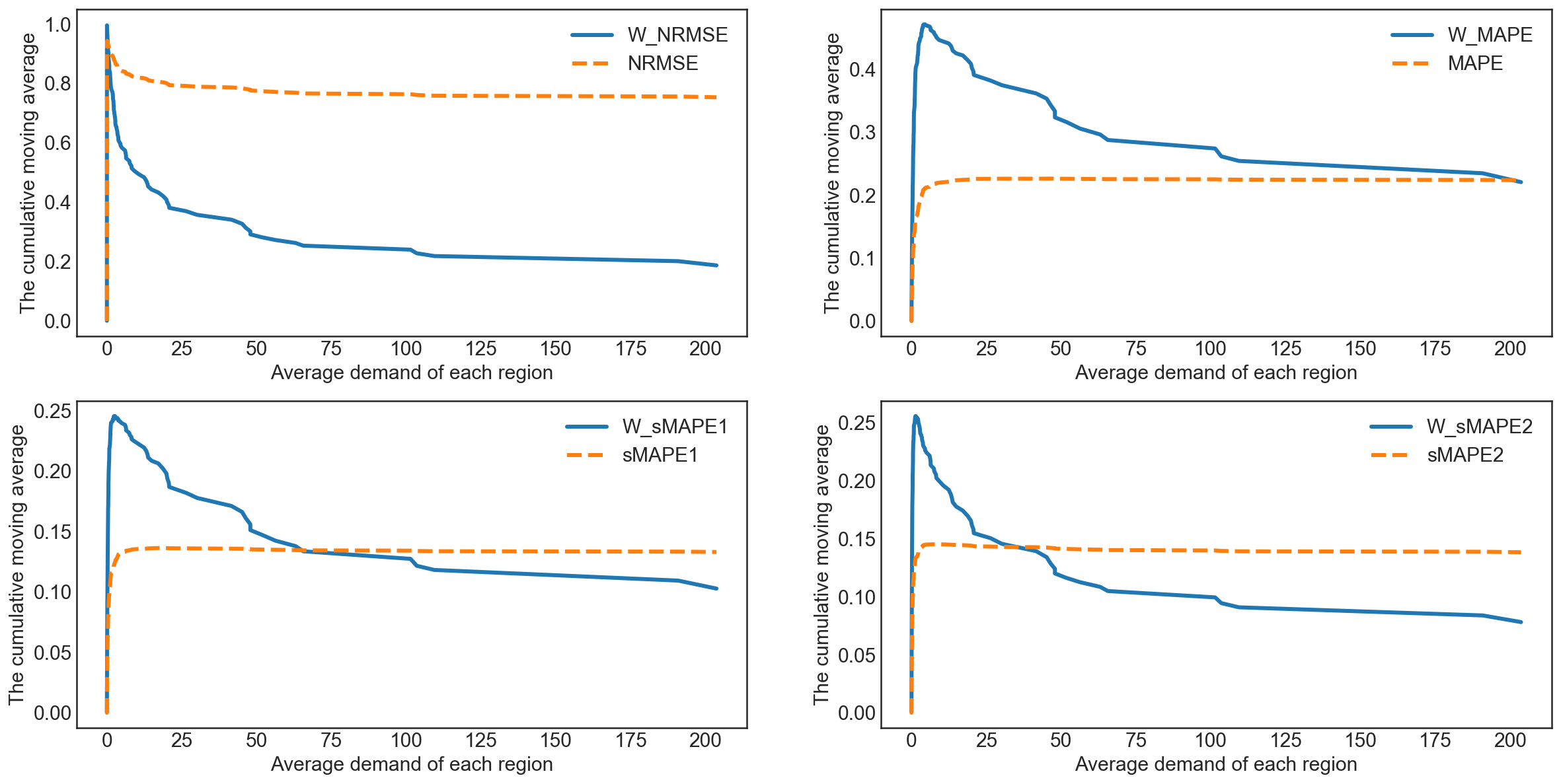}
    \caption{Comparison of the weighted and non-weighted evaluation functions.}
    \label{weighted}
\end{figure}

Figure \ref{weighted} shows the difference between the two kinds of metrics. It is obvious that the regions with higher demands now contribute more in the weighted-arithmetic-based metrics, which might be more preferable for the ride-sourcing platform, since higher priority should be given to regions with higher demand to guarantee the system efficiency. Moreover, the weighted arithmetic is less sensitive to the grid size as shown in Table \ref{Grid2}.

The prediction errors calculated based on weighted-arithmetic metrics are shown in Table \ref{weighted_label}. Now the predictive performance rankings of the models under most metrics are consistent, except for W$_{-}$MAPE. Under weighted-arithmetic metrics, the LC-ST-FCN proposed in this study outperforms all the other benchmark mondels in all the five metrics. There are some interesting observations below:

i) In terms of predictive performance,  LC-ST-FCN $>$ LC-FCN $>$ FCN, which indicates that 3D convolutions are more suitable for spatiotemporal feature learning compared to 2D convolutions; And the locally connected convolutional layers can deal with the impact of local statistical differences well, better than standard convolutional layers. The LC-ST-FCN, LC-FCN and FCN are the top three ranked models. That may due to the fully convolutional architecture of them, which can help maintain the spatial coordinates of the input and avoid losing spatial information between layers.

ii) In terms of predictive performance, FCN $>$ CNN. The only difference between the two structures is the last two layers. FCN and CNN use standard convolution layers, and fully connected layers, respectively. In our experiments, the ratio between the total numbers of parameters of the last two layers for the two models are nearly 1:8000. One of the reasons why FCN is superior to CNN might be that the fully connected layer in CNN is affected by a very large patch of the input and learning the optimal combination of parameters for numerous neurons is difficult.

iii) In terms of predictive performance, LC-ST-FCN$>$LC-ST-FCN (diff). The good predictive performance of additive model demonstrates that the trend and periodicity of demand data play an important role in demand forecasting. Both LC-ST-FCN (diff) and LC-ST-FCN can learn from these information, yet the results show that the 3D convolution operation is more powerful in feature extraction than difference method.

In order to further compare the predictive performance of each model in regions with different levels of uncertainty, we calculated the evaluation metrics for both $G_1$ and $G_2$ ( Table \ref{G_12}). It can be observed that for all the models, the weighted percentage errors for $G_2$ are nearly doubled, which indicates that uncertainties in the time series data can greatly reduce the models' prediction ability.

\begin{table}[ht]
    \small
    \centering
    \caption{\newline Prediction error of the LC-ST-FCN model for different grid sizes using the weighted method.}
    \label{Grid2}
\begin{tabular}{@{}llllll@{}}
\toprule
Grid size      & RMSE & W$_{-}$NRMSE (\%) & W$_{-}$MAPE (\%) & W$_{-}$sMAPE1 (\%) & W$_{-}$sMAPE2 (\%) \\ \midrule
$16\times16$     & 1.67 & 18.76         & 22.09        & 10.28          & 7.83           \\
$64\times64$     & 0.35 & 47.04         & 37.16        & 18.53          & 17.63          \\ \bottomrule
\end{tabular}
\end{table}

\begin{table}[ht]
    \footnotesize
    \centering
    \caption{\newline Predictive performance comparison (weighted).}
    \label{weighted_label}
\begin{tabular}{@{}llllll@{}}
\toprule
Model            & RMSE     & W$_{-}$NRMSE (\%) & W$_{-}$MAPE (\%) & W$_{-}$sMAPE1 (\%) & W$_{-}$sMAPE2 (\%) \\ \midrule
LC-ST-FCN        & \underline{1.67 (1)} & \underline{18.76 (1)}     & \underline{22.09 (1)}    & \underline{10.28 (1)}      & \underline{7.83 (1)}       \\
LC-FCN           & 1.69 (2) & 19.09 (2)     & 23.39 (2)    & 10.61 (2)      & 7.96 (2)       \\
FCN              & 1.78 (3) & 19.80 (3)     & 23.84 (3)    & 10.88 (3)      & 8.24 (3)       \\
CNN              & 1.82 (4) & 20.41 (4)     & 24.43 (5)    & 11.02 (4)      & 8.45 (4)       \\
Additive model   & 1.86 (5) & 20.77 (5)     & 26.99 (6)    & 11.56 (5)      & 8.63 (5)       \\
Conv-LSTM        & 1.95 (6) & 21.61 (6)     & 24.32 (4)    & 11.95 (6)      & 9.16 (6)       \\
ANN              & 2.04 (7) & 23.42 (7)     & 28.57 (8)    & 12.79 (7)      & 9.82 (7)       \\
LC-ST-FCN (diff) & 2.17 (8) & 24.32 (8)     & 27.02 (7)    & 12.92 (8)      & 9.96 (8)       \\ \bottomrule
\end{tabular}
\end{table}

\begin{table}[ht]
    \footnotesize
    \centering
    \caption{\newline Predictive performance of $G_1$ and $G_2$ (weighted method).}
    \label{G_12}
\begin{tabular}{@{}llllll@{}}
\toprule
Model            & RMSE      & W$_{-}$NRMSE (\%) & W$_{-}$MAPE (\%)  & W$_{-}$sMAPE1 (\%) & W$_{-}$sMAPE2 (\%) \\ \midrule
\multicolumn{6}{c}{regions ($G_1/G_2$)}                                                         \\ \midrule
LC-ST-FCN        & 4.35/0.92 & 14.40/37.21   & 18.38/37.79   & 8.45/18.04     & 6.12/15.05     \\
LC-FCN           & 4.47/0.91 & 14.82/37.14   & 19.71/38.91   & 8.80/18.23     & 6.29/15.04     \\
FCN              & 4.57/1.01 & 15.09/39.72   & 19.77/41.01 & 8.95/19.04   & 6.42/15.93   \\
CNN              & 4.83/0.98 & 16.01/39.01   & 20.76/39.99   & 9.24/18.57     & 6.76/15.63     \\
Additive model   & 4.75/1.05 & 15.68/42.29   & 23.21/42.99   & 9.69/19.51     & 6.72/16.7      \\
Conv-LSTM        & 5.14/1.06 & 16.97/41.23   & 20.72/39.55   & 10.15/19.59    & 7.36/16.77     \\
ANN              & 5.79/0.99 & 19.35/40.65   & 24.64/45.18   & 11.06/20.07    & 8.19/16.7      \\
LC-ST-FCN (diff) & 5.60/1.21  & 18.51/48.92   & 22.78/44.99   & 10.98/21.15    & 7.85/18.9      \\
ARIMA            & 5.52/-    & 18.23/-       & 22.09/-       & 10.75/-        & 7.74/-         \\ \bottomrule
\end{tabular}
\end{table}

\newpage
\section{Conclusions}
\label{S:5}

CNN has been demonstrated as an effective tool for learning information with spatial structure, which has led to breakthroughs in almost all machine learning tasks. In this paper, a fusion convolutional model, LC-ST-FCN, is established for demand forecasting of ride-sourcing services. New model structure and evaluation metrics are developed to deal with the unique characteristics of ride-sourcing services. A real dataset from DiDiChuxing platform is used for model evaluation and comparison. In the experiments, our model outperforms all the benchmark models in terms of all the weighted-arithmetic-based metrics and most of the arithmetic-mean-based metrics. The weighted-arithmetic-based metrics show better consistency in performance evaluation than the arithmetic-mean-based metrics, because they take demand level and unbalanced demand distribution into consideration. Moreover, we show that prediction results can be greatly affected by the local statistics, since in our experiments, the percentage errors in regions with high uncertainty are nearly twice of those in regions with low uncertainty. In this paper, the training data of our proposed model is generated by grid-based partition, which is carried out independently of the model training. In the future work, we expect to explore a learnable partition algorithm which has the ability of automatic region partition by learning the demand statistics of ride-sourcing service platforms.

\section*{Acknowledgement}
The dataset used in this paper comes from DiDiChuxing. The work described in this paper was supported by the National Natural Science Foundation of China (71622007, 71861167001) and the National Key Research and Development Program of China (2018YFB1600902).






\bibliographystyle{elsarticle-harv.bst}
\bibliography{reference_402}

\begin{thebibliography}{37}
\expandafter\ifx\csname natexlab\endcsname\relax\def\natexlab#1{#1}\fi
\expandafter\ifx\csname url\endcsname\relax
  \def\url#1{\texttt{#1}}\fi
\expandafter\ifx\csname urlprefix\endcsname\relax\def\urlprefix{URL }\fi

\bibitem[{Auer et~al.(2017)Auer, Rehborn, Molzahn, and
  Koller}]{auer2017traffic}
Auer, M., Rehborn, H., Molzahn, S.-E., Koller, M., 2017. Traffic services for
  vehicles: the process from receiving raw probe data to space-time diagrams
  and the resulting traffic service. Frontiers of Engineering Management.
  4~(4), 490--497.

\bibitem[{Brockwell and Davis(2015)}]{RN61}
Brockwell, P.~J., Davis, R.~A., 2015. Time Series: Theory and Methods.
  Springer-Verlag.

\bibitem[{Chang et~al.(2009)Chang, Tai, and Hsu}]{RN22}
Chang, H.~W., Tai, Y.~C., Hsu, Y.~J., 2009. Context-aware taxi demand hotspots
  prediction. International Journal of Business Intelligence and Data Mining.
  5~(1), 3--18.

\bibitem[{Chen and Li(2018)}]{RN51}
Chen, Xiqun~Michael, C. C. L.~N., Li, L., 2018. Spatial visitation prediction
  of on-demand ride services using the scaling law. Physica A: Statistical
  Mechanics and its Applications. 508, 84--94.

\bibitem[{Chen et~al.(2017{\natexlab{a}})Chen, Zahiri, and Zhang}]{RN37}
Chen, X., Zahiri, M., Zhang, S., 2017{\natexlab{a}}. Understanding
  ridesplitting behavior of on-demand ride services: An ensemble learning
  approach. Transportation Research Part C: Emerging Technologies. 76, 51--70.

\bibitem[{Chen et~al.(2017{\natexlab{b}})Chen, Chen, Zheng, and
  Chen}]{chen2017understanding}
Chen, X.~M., Chen, X., Zheng, H., Chen, C., 2017{\natexlab{b}}. Understanding
  network travel time reliability with on-demand ride service data. Frontiers
  of Engineering Management. 4~(4), 388--398.

\bibitem[{Deng and Ji(2011)}]{RN33}
Deng, Z., Ji, M., 2011. Spatiotemporal structure of taxi services in shanghai:
  Using exploratory spatial data analysis. In: 2011 19th International
  Conference on Geoinformatics. pp. 1--5.

\bibitem[{Dong et~al.(2018)Dong, Wang, Li, and Zhang}]{RN1}
Dong, Y., Wang, S., Li, L., Zhang, Z., 2018. An empirical study on travel
  patterns of internet based ride-sharing. Transportation Research Part C:
  Emerging Technologies. 86, 1--22.

\bibitem[{Gregor and Lecun(2010)}]{RN67}
Gregor, K., Lecun, Y., 2010. Emergence of complex-like cells in a temporal
  product network with local receptive fields. arXiv:1006.0448.

\bibitem[{Hara et~al.(2017)Hara, Kataoka, and Satoh}]{RN72}
Hara, K., Kataoka, H., Satoh, Y., 2017. Learning spatio-temporal features with
  3d residual networks for action recognition. In: Proceedings of the IEEE
  international conference on computer vision. pp. 3154--3160.

\bibitem[{Huang et~al.(2012)Huang, Lee, and Learned-Miller}]{RN68}
Huang, G.~B., Lee, H., Learned-Miller, E., 2012. Learning hierarchical
  representations for face verification with convolutional deep belief
  networks. In: 2012 IEEE Conference on Computer Vision and Pattern
  Recognition. pp. 2518--2525.

\bibitem[{Huang et~al.(2014)Huang, Song, Hong, and Xie}]{RN35}
Huang, W., Song, G., Hong, H., Xie, K., 2014. Deep architecture for traffic
  flow prediction: Deep belief networks with multitask learning. IEEE:
  Transactions on Intelligent Transportation Systems. 15~(5), 2191--2201.

\bibitem[{Jessica~Christian(2018)}]{RN62}
Jessica~Christian, S.~E., 2018. Study: Half of sf’s increase in traffic
  congestion due to uber, lyft.
  http://www.sfexaminer.com/study-half-sfs-increase-traffic-congestion-due-uber-lyft/.

\bibitem[{Kaltenbrunner et~al.(2010)Kaltenbrunner, Meza, Grivolla, Codina, and
  Banchs}]{RN21}
Kaltenbrunner, A., Meza, R., Grivolla, J., Codina, J., Banchs, R., 2010. Urban
  cycles and mobility patterns: Exploring and predicting trends in a
  bicycle-based public transport system. Pervasive and Mobile Computing. 6~(4),
  455--466.

\bibitem[{Karpathy et~al.(2014)Karpathy, Toderici, Shetty, Leung, Sukthankar,
  and Fei-Fei}]{RN75}
Karpathy, A., Toderici, G., Shetty, S., Leung, T., Sukthankar, R., Fei-Fei, L.,
  2014. Large-scale video classification with convolutional neural networks.
  In: Proceedings of the IEEE conference on Computer Vision and Pattern
  Recognition. pp. 1725--1732.

\bibitem[{Ke et~al.(2017)Ke, Zheng, Yang, and Chen}]{RN24}
Ke, J., Zheng, H., Yang, H., Chen, X., 2017. Short-term forecasting of
  passenger demand under on-demand ride services: A spatio-temporal deep
  learning approach. Transportation Research Part C: Emerging Technologies. 85,
  591--608.

\bibitem[{Krizhevsky(2012)}]{RN18}
Krizhevsky, A., S. I. H.~G., 2012. Imagenet classification with deep
  convolutional neural networks. In: Advances in Neural Information Processing
  Systems. pp. 1097--1105.

\bibitem[{Kullback and Leibler(1951)}]{RN71}
Kullback, S., Leibler, R.~A., 1951. On information and sufficiency. Annals of
  Mathematical Statistics. 22~(1), 79--86.

\bibitem[{Lecun et~al.(2015)Lecun, Bengio, and Hinton}]{RN34}
Lecun, Y., Bengio, Y., Hinton, G., 2015. Deep learning. Nature. 521~(7553),
  436.

\bibitem[{LeCun et~al.(1989)LeCun, Boser, Denker, Henderson, Howard, Hubbard,
  and Jackel}]{RN41}
LeCun, Y., Boser, B., Denker, J.~S., Henderson, D., Howard, R.~E., Hubbard, W.,
  Jackel, L.~D., 1989. Backpropagation applied to handwritten zip code
  recognition. Neural computation. 1~(4), 541--551.

\bibitem[{Ma et~al.(2017)Ma, Dai, He, Ma, Wang, and Wang}]{RN83}
Ma, X., Dai, Z., He, Z., Ma, J., Wang, Y., Wang, Y., 2017. Learning traffic as
  images: a deep convolutional neural network for large-scale transportation
  network speed prediction. Sensors. 17~(4), 818.

\bibitem[{Ma et~al.(2015)Ma, Yu, Wang, and Wang}]{RN38}
Ma, X., Yu, H., Wang, Y., Wang, Y., 2015. Large-scale transportation network
  congestion evolution prediction using deep learning theory. Plos One. 10~(3),
  e0119044.

\bibitem[{Moreira-Matias et~al.(2013)Moreira-Matias, Gama, Ferreira,
  Mendes-Moreira, and Damas}]{RN15}
Moreira-Matias, L., Gama, J., Ferreira, M., Mendes-Moreira, J., Damas, L.,
  2013. Predicting taxi–passenger demand using streaming data. IEEE:
  Transactions on Intelligent Transportation Systems. 14~(3), 1393--1402.

\bibitem[{Shelhamer et~al.(2014)Shelhamer, Long, and Darrell}]{RN17}
Shelhamer, E., Long, J., Darrell, T., 2014. Fully convolutional networks for
  semantic segmentation. IEEE: Transactions on Pattern Analysis and Machine
  Intelligence. 39~(4), 640--651.

\bibitem[{Shuiwang et~al.(2013)Shuiwang, Ming, and Kai}]{RN63}
Shuiwang, J., Ming, Y., Kai, Y., 2013. 3d convolutional neural networks for
  human action recognition. IEEE: Transactions on Pattern Analysis and Machine
  Intelligence. 35~(1), 221--231.

\bibitem[{Simonyan and Zisserman(2014{\natexlab{a}})}]{RN76}
Simonyan, K., Zisserman, A., 2014{\natexlab{a}}. Two-stream convolutional
  networks for action recognition in videos. pp. 568--576.

\bibitem[{Simonyan and Zisserman(2014{\natexlab{b}})}]{RN45}
Simonyan, K., Zisserman, A., 2014{\natexlab{b}}. Very deep convolutional
  networks for large-scale image recognition. arXiv:1409.1556.

\bibitem[{Szegedy et~al.(2015)Szegedy, Liu, Jia, Sermanet, Reed, Anguelov,
  Erhan, Vanhoucke, and Rabinovich}]{RN44}
Szegedy, C., Liu, W., Jia, Y., Sermanet, P., Reed, S., Anguelov, D., Erhan, D.,
  Vanhoucke, V., Rabinovich, A., 2015. Going deeper with convolutions. In:
  Proceedings of the IEEE conference on computer vision and pattern
  recognition. pp. 1--9.

\bibitem[{Taigman et~al.(2014)Taigman, Yang, Ranzato, and Wolf}]{RN66}
Taigman, Y., Yang, M., Ranzato, M., Wolf, L., 2014. Deepface: Closing the gap
  to human-level performance in face verification. In: Proceedings of the IEEE
  conference on computer vision and pattern recognition. pp. 1701--1708.

\bibitem[{Tran et~al.(2015)Tran, Bourdev, Fergus, Torresani, and Paluri}]{RN73}
Tran, D., Bourdev, L., Fergus, R., Torresani, L., Paluri, M., 2015. Learning
  spatiotemporal features with 3d convolutional networks. In: Proceedings of
  the IEEE international conference on computer vision. pp. 4489--4497.

\bibitem[{Vazifeh et~al.(2018)Vazifeh, Santi, Resta, Strogatz, and Ratti}]{RN4}
Vazifeh, M.~M., Santi, P., Resta, G., Strogatz, S.~H., Ratti, C., 2018.
  Addressing the minimum fleet problem in on-demand urban mobility. Nature.
  557~(7706), 534--538.

\bibitem[{Xingjian and Woo(2015)}]{RN60}
Xingjian, S.H.I., C. Z. W. H. Y. D. W.~W., Woo, W., 2015. Convolutional lstm
  network: A machine learning approach for precipitation nowcasting. In:
  Advances in neural information processing systems. pp. 802--810.

\bibitem[{Yao et~al.(2018)Yao, Wu, Ke, Tang, Jia, Lu, Gong, Ye, and Li}]{RN84}
Yao, H., Wu, F., Ke, J., Tang, X., Jia, Y., Lu, S., Gong, P., Ye, J., Li, Z.,
  2018. Deep multi-view spatial-temporal network for taxi demand prediction.
  In: Thirty-Second AAAI Conference on Artificial Intelligence.

\bibitem[{Yu et~al.(2017)Yu, Li, Shahabi, Demiryurek, and Liu}]{RN78}
Yu, R., Li, Y., Shahabi, C., Demiryurek, U., Liu, Y., 2017. Deep learning: A
  generic approach for extreme condition traffic forecasting. In: Proceedings
  of the 2017 SIAM International Conference on Data Mining. pp. 777--785.

\bibitem[{Yuan et~al.(2011)Yuan, Zheng, Zhang, Xie, and Sun}]{RN23}
Yuan, J., Zheng, Y., Zhang, L., Xie, X., Sun, G., 2011. Where to find my next
  passenger. In: Proceedings of the 13th international conference on Ubiquitous
  computing. pp. 109--118.

\bibitem[{Zhang et~al.(2017)Zhang, Zheng, and Qi}]{RN82}
Zhang, J., Zheng, Y., Qi, D., 2017. Deep spatio-temporal residual networks for
  citywide crowd flows prediction. In: Thirty-First AAAI Conference on
  Artificial Intelligence.

\bibitem[{Zhao et~al.(2016)Zhao, Khryashchev, Freire, Silva, and Vo}]{RN32}
Zhao, K., Khryashchev, D., Freire, J., Silva, C., Vo, H., 2016. Predicting taxi
  demand at high spatial resolution: Approaching the limit of predictability.
  In: 2016 IEEE International Conference on Big Data (Big Data). pp. 833--842.

\end{thebibliography}







\end{document}